\documentclass[letterpaper, 10 pt, conference]{ieeeconf}  

\IEEEoverridecommandlockouts                              

\title{\LARGE\bf
From Vision to Assistance: Gaze and Vision-Enabled Adaptive Control for a Back-Support Exoskeleton
}

\author{
Alessandro Leanza$^{1,*}$\thanks{*Corresponding author: alessandro.leanza@supsi.ch},
Paolo Franceschi$^{1}$,
Blerina Spahiu$^{2}$,
Loris Roveda$^{1,3}$%
\thanks{$^{1}$Department of Innovative Technologies (DTI), Dalle Molle Institute for Artificial Intelligence (IDSIA),
University of Applied Sciences and Arts of Southern Switzerland (SUPSI),
6900 Lugano, Switzerland.}%
\thanks{$^{2}$Department of Informatics, Systems and Communication (DISCo),
University of Milano-Bicocca, 20126 Milano, Italy.}%
\thanks{$^{3}$Department of Mechanical Engineering,
Politecnico di Milano, 20156 Milano, Italy.}%
}

\usepackage{url}

\usepackage{amsmath,amsfonts,amstext}
\usepackage{algorithmic}
\usepackage{algorithm}
\usepackage{array}
\usepackage{textcomp}
\usepackage{stfloats}
\usepackage{verbatim}
\usepackage{hyperref}
\usepackage{mathptmx} 
\usepackage{amssymb}  
\usepackage{graphicx}
\usepackage{tabularx}
\usepackage{epstopdf}
\usepackage{booktabs}
\usepackage{array}
\usepackage{bm}
\usepackage{calc}
\usepackage{multicol}
\usepackage{pslatex}
\usepackage{caption, subcaption}
\usepackage{multirow}
\usepackage{siunitx}
\usepackage{caption}
\usepackage{xcolor}

\usepackage{tikz}
\usetikzlibrary{arrows.meta,positioning,fit,calc}

\usepackage[dvipsnames]{xcolor} 

\newcommand{\dg}[1]{\textcolor{ForestGreen}{#1}}   
\newcommand{\dy}[1]{\textcolor{YellowOrange}{#1}}     
\newcommand{\db}[1]{\textcolor{NavyBlue}{#1}}      

\begin{document}

\newcommand{\boldsymbold}[1]{\dot{\boldsymbol{#1}}} 
\newcommand{\boldsymboldd}[1]{\ddot{\boldsymbol{#1}}}

\newcommand{\Brobot}{\mathbf{B}(\mathbf{q})}
\newcommand{\Crobot}{\mathbf{C}(\mathbf{q},\mathbf{\dot{q}})}
\newcommand{\Rfriction}{\boldsymbol{\tau}_f(\mathbf{q},\dot{\mathbf{q}})}
\newcommand{\grobot}{\mathbf{g}(\mathbf{q})}
\newcommand{\Jrobot}{\mathbf{J}(\mathbf{q})}
\newcommand{\Jrobott}{\mathbf{J}(\mathbf{q}(k_t))}
\newcommand{\dJrobot}{\mathbf{\dot{J}}(\mathbf{q},\mathbf{\dot{q}})}
\newcommand{\hrobot}{\mathbf{h}_{ext}}
\newcommand{\taurobot}{\boldsymbol{\tau}}
\newcommand{\qrobot}{\mathbf{q}}
\newcommand{\dqrobot}{\dot{\mathbf{q}}}
\newcommand{\ddqrobot}{\ddot{\mathbf{q}}}
\newcommand{\Fparam}{\mathbf{F}_p}
\newcommand{\Fparamold}{\mathbf{F}_{p,old}}
\newcommand{\hf}{\mathbf{h}_{f,c}}
\newcommand{\tauf}{\boldsymbol{\tau}_{f}}

\newcommand{\ie}{\textit{i.e.}}
\newcommand{\eg}{\textit{e.g.}}
\newcommand{\etc}{\textit{etc}}
\newcommand{\wrt}{w.r.t. }
\newcommand{\ma}[1]{\mathbf{#1}}        
\newcommand{\vt}[1]{\mathbf{#1}}        
\newcommand{\vg}[1]{\boldsymbol{#1} }        
\newcommand{\vtd}[1]{\dot{\mathbf{#1}}}        
\newcommand{\vgd}[1]{\dot{\boldsymbol{#1}}}        
\newcommand{\vtdd}[1]{\ddot{\mathbf{#1}}}        
\newcommand{\vgdd}[1]{\ddot{\boldsymbol{#1}}}        
\newcommand{\sizeotto}{\fontsize{8pt}{8pt}\selectfont}
\newcommand{\sizenove}{\fontsize{9pt}{9pt}\selectfont}
\newcommand{\sizedieci}{\fontsize{10pt}{10pt}\selectfont}
\newcommand{\dd}{\mathrm{d}}

\newcounter{CntRemark} 
\newcommand{\newremark}{\addtocounter{CntRemark}{1}\noindent\textbf{Remark \theCntRemark. }}

\newcommand{\red}[1]{\textcolor{red}{#1}}

\maketitle
\thispagestyle{empty}
\pagestyle{empty}

\begin{abstract}
Back-support exoskeletons have been proposed to mitigate spinal loading in industrial handling, yet their effectiveness critically depends on timely and context-aware assistance. 
Most existing approaches rely either on load-estimation techniques (\textit{e.g.}, EMG, IMU) or on vision systems that do not directly inform control. 
In this work, we present a vision-gated control framework for an active lumbar occupational exoskeleton that leverages egocentric vision with wearable gaze tracking. 
The proposed system integrates real-time grasp detection from a first-person YOLO-based perception system, a finite-state machine (FSM) for task progression, and a variable admittance controller to adapt torque delivery to both posture and object state. 
A user study with 15 participants performing stooping load lifting trials under three conditions (no exoskeleton, exoskeleton without vision, exoskeleton with vision) shows that vision-gated assistance significantly reduces perceived physical demand and improves fluency, trust, and comfort. 
Quantitative analysis reveals earlier and stronger assistance when vision is enabled, while questionnaire results confirm user preference for the vision-gated mode. 
These findings highlight the potential of egocentric vision to enhance the responsiveness, ergonomics, safety, and acceptance of back-support exoskeletons.
\end{abstract}


\section{Introduction}

Back pain is one of the most reported injuries among the work-related disorders \cite{article2-Ibrahim}. 
More than 40\% of the EU workers are suffering from this pathology \cite{article2-naf}.
To overcome this issue, solutions such as cranes, wearable devices, and collaborative robots have been developed \cite{article2}. 
Among the others, back-support exoskeletons \cite{ali2021systematic} redistribute the load on the legs and decrease the load on the spine, while remaining wearable and portable, thus providing effective support also in dynamic and unstructured handling scenarios, unlike cranes or collaborative robots, which are mainly suitable for repetitive and structured tasks.

Two types of industrial exoskeletons for back support are studied in the literature: passive and active exoskeletons, which can both be characterized by a soft \cite{bianchi2024friction} or rigid \cite{article25} design.
Passive devices rely on mechanical components such as springs and dampers to provide assistive forces without the need for an actuation system \cite{article1}. 
Examples include squat-lifting aids \cite{article2-wehner2009lower} and flexible beam designs parallel to the spine \cite{article2-naf}, which increase sagittal range of motion and reduce muscle load. 
Passive solutions remain more common in industry, both in adoption and validation studies \cite{article10}, but they can only generate restorative forces that guide the body toward equilibrium, lacking adaptability to varying task demands.  
In contrast, active exoskeletons can modulate support in a task-- and user--dependent manner, adapting assistance to the task phase, user characteristics, and personal preferences \cite{article2}. However, such devices are heavier and more expensive (due to electronics, motors, and batteries) in comparison to passive ones, and currently available controllers are still not intelligent enough to provide a real practical advantage.

\begin{figure}[b!]
    \centering
    \includegraphics[width=0.9\columnwidth]{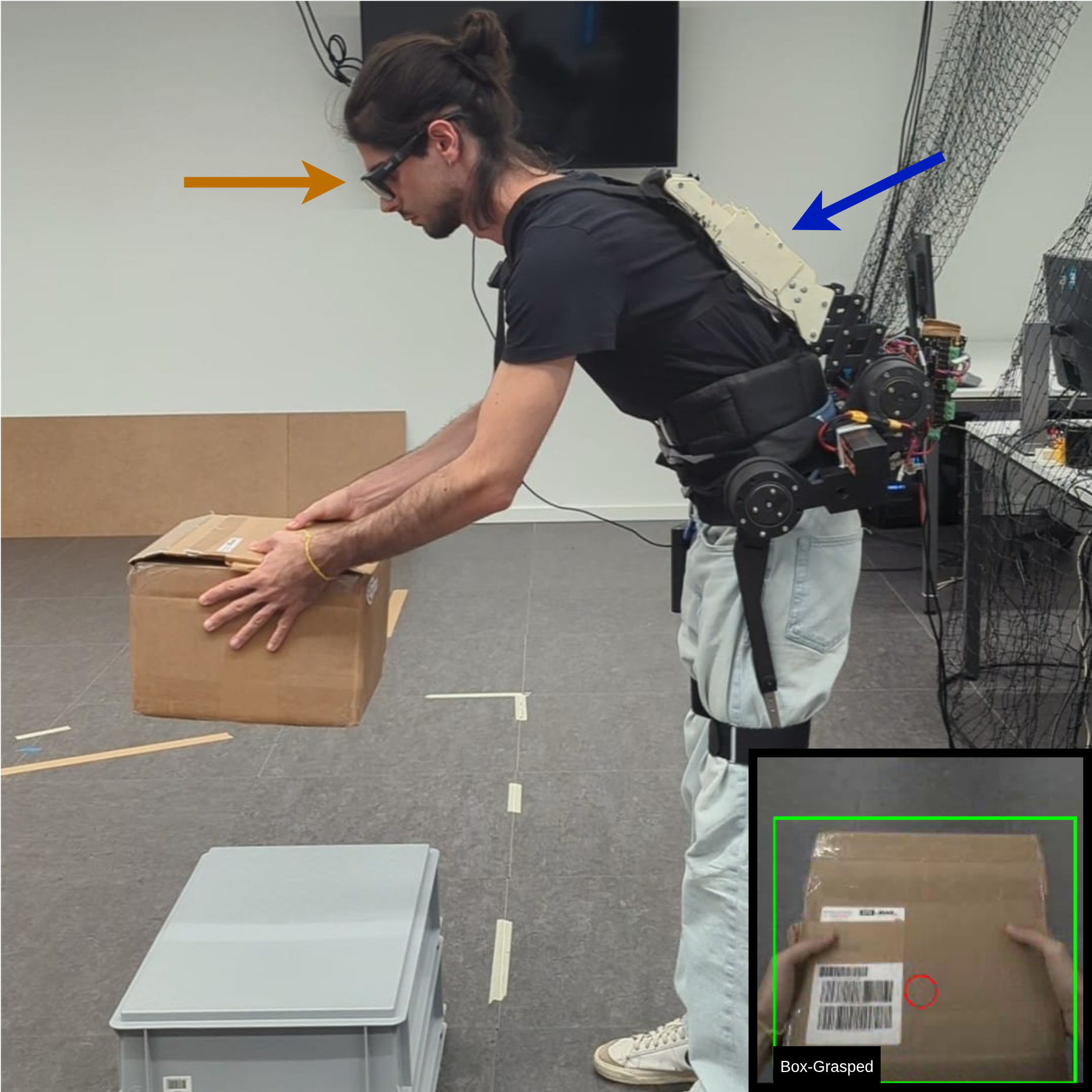}
    \caption{User performing the lifting task with the back–support exoskeleton (\db{blue arrow}). The inset (bottom right) provides the egocentric perspective from the gaze–tracking glasses (\dy{orange arrow}), where the target box is highlighted in \dg{green box} and the user’s gaze fixation is marked in \red{red circle}.}

    \label{fig:activeexo}
\end{figure}

To promote the adoption of active occupational exoskeletons, which are more proactive and adaptive to the operational context and user, this paper presents a novel vision-gated control framework for an active lumbar occupational exoskeleton. Our system integrates gaze-aware grasp detection, a finite-state machine policy, and a variable admittance controller to deliver adaptive support during stooping tasks (Figure~\ref{fig:activeexo}). 
To assess its performance, the proposed control framework is evaluated in a controlled user study with 15 participants performing repeated box-handling trials by means of questionnaires comparing three testing conditions: (i) no exoskeleton, (ii) exoskeleton without vision, and (iii) exoskeleton with vision.

\subsection{Related Work}


To be effective in industrial handling (\ie, lifting and putting down objects), an active occupational exoskeleton must perceive the ongoing situation and adapt the user assistance accordingly.

For the implementation of active assistance, classical control approaches (both model-based or data-driven) rely on impedance and admittance control \cite{anam2012active} to ensure compliance, and user-related signals only are used by the controller. 
Such approaches include user-applied torque-based controllers \cite{article1}, IMU-based controllers \cite{koopman2019effect}, and EMG-driven controllers \cite{zhang2025review}. 
For these methodologies, some limitations can be highlighted. Invasive and user-dependent sensors (\eg, EMGs) are needed for the controller. Furthermore, manual tuning/controller selection affects the performance of the human-exoskeleton collaboration. Finally, such controllers are purely reactive and blind with respect to the operative scenario. 
Indeed, by means of standard techniques, the exoskeleton is not capable of perceiving the operative environment, thus not being able to provide any adaptive assistance.

To address such issues, vision-based control strategies have been proposed. Visual information allows the exoskeleton to modulate its assistance to the user based on the current scenario. 
This approach is widely used in lower limb exoskeletons, particularly to detect walking phases \cite{wang2024review} and terrains \cite{laschowski2020exonet}.
In \cite{tricomi2023environment}, the amplitude
of walking assistance is modulated depending on the environment, classified by a vision module.
In the context of empowering exoskeletons aimed at reducing human effort, vision has been employed to estimate load weight and integrate this information into the control loop.
\cite{10333973} proposes a computer vision layer for external weight estimation, to offer collaborative assistance for an upper limb exoskeleton, with images recorded with a chest-mounted camera.
Also \cite{prete2025computer} uses an embedded camera with a YOLO-based estimation for the payload in an active back-support exoskeleton, classifying it as low, medium, or heavy, with limited consideration of its integration in the control loop and no experimental evaluation.

In this direction, egocentric vision and gaze-based interaction can be effective to improve the human-exoskeleton collaboration, resulting in a natural and effective communication modality. 
Such approaches have proven highly effective in domains such as prosthetics, where gaze cues improve grasp classification \cite{cognolato2022improving}, assistive robotics, where gaze-camera systems predict user intentions before movement onset \cite{he2025gaze}, and collaborative robotics \cite{shahid2025gear}, where gaze is used to interact with a cobot for assisting the user in a dual assembly task. 
Yet, despite this promise, egocentric vision has not been explored in back-support exoskeletons, where the correct timing of torque delivery is crucial.

Addressing this gap is significant because usability and trust are critical for industrial adoption. 
Field studies show that workers’ intention to use exoskeletons depends strongly on comfort, timing of support, and perceived effectiveness \cite{Hess2025}. 
Mistimed or intrusive assistance reduces trust and discourages long-term use, while effective, well-timed relief increases acceptance.

\subsection{Paper Contribution} 

Based on the state-of-the-art (SotA) analysis discussed above, the main contributions of this work are:
\begin{itemize}
    \item We introduce an egocentric vision module combining gaze tracking and object detection to robustly identify when the user grasps a load, enabling context-aware assistance for an active occupational exoskeleton.
    \item We design a variable admittance control policy gated by the vision module, which modulates assistance based on both posture and grasp state, improving responsiveness and safety.
    \item We conduct a user study with 15 healthy volunteers comparing three conditions (no exoskeleton, exoskeleton without vision, exoskeleton with vision), evidence that vision-gated assistance reduces perceived physical demand and increases fluency, trust, and comfort in the active occupational exoskeleton.
\end{itemize}


\section{System Modeling}
\label{sec:exo}

\begin{figure}[b!]
    \centering
    \includegraphics[width=0.85\columnwidth]{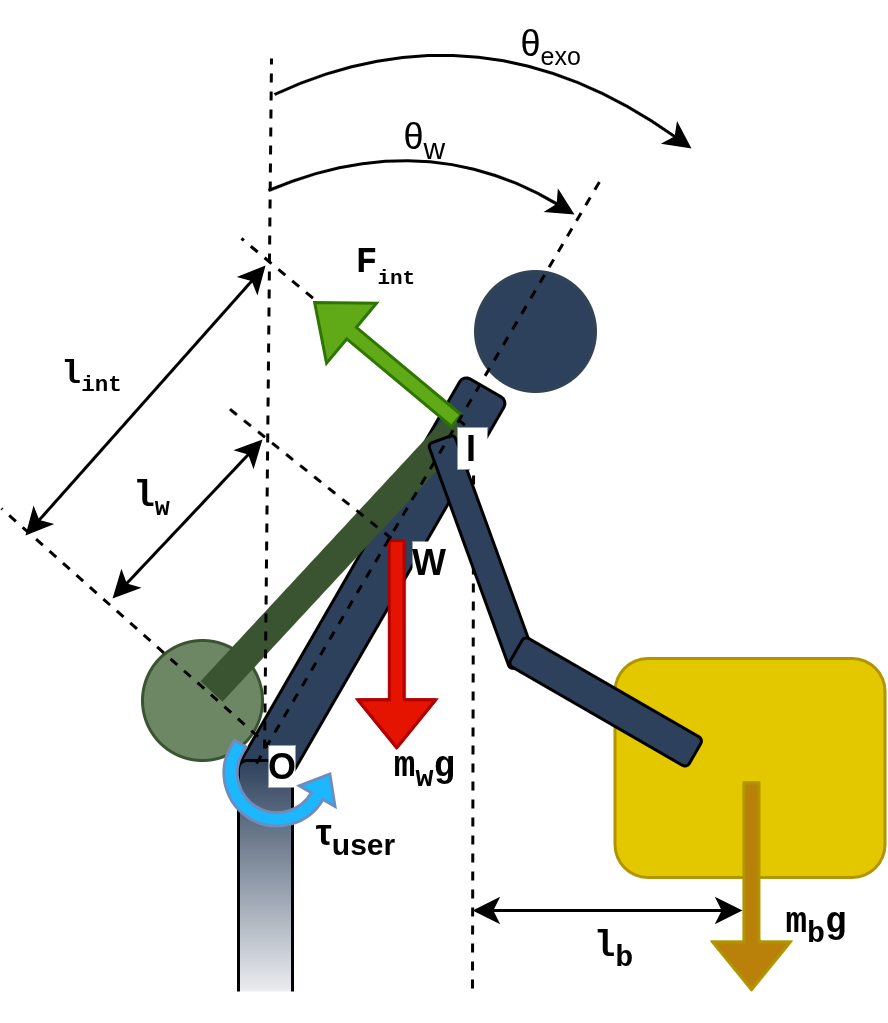}
    \caption{System model. The forces and torques acting on the exoskeleton and the user during a handling task. 
    }
    \label{fig:forze-coppie}
\end{figure}

For the active occupational lumbar exoskeleton implementation, we adopt a SotA backbone-based kinematics \cite{article2,article3} and a 3D-printed cycloidal actuation system \cite{article25}. The exoskeleton is sensorless, \textit{i.e.}, no torque/force sensors are embedded in its design.

For the human-exoskeleton system modeling in Figure~\ref{fig:activeexo}, we analyze the quasi-static moment balance of the human–exoskeleton system during object handling (Figure~\ref{fig:forze-coppie}).
Let $m_w$ and $m_b$ denote the masses of the upper body and of the handled box, respectively. Their weights are $m_w g$ and $m_b g$, with $g$ the gravitational acceleration.
Let $\theta_w$ be the trunk inclination with respect to the vertical, $\ell_w{=}OW$ the distance from hip $O$ to the upper body center of mass location $W$, and $\ell_{\text{int}}{=}OI$ the distance from $O$ to the interaction point $I$ on the upper back. 
The box center of mass lies at a horizontal offset $\ell_b(\theta_w)$ forward of $I$\footnote{In practice, $\ell_b$ is approximated as constant, since variations across the tested range of trunk inclinations are small.}.
The human-exoskeleton connection pad is assumed normal to the back, so the interaction force $F_{\text{int}}$ at $I$ is orthogonal to the trunk.

The gravitational torques at the hip joint $O$ are:
\begin{align}
\tau_w(\theta_w) &= m_w g\, \ell_w \sin\theta_w, \label{eq:tau_w}\\
\tau_{\text{box}}(\theta_w) &= m_b g\,\bigl(x_I(\theta_w)+\ell_b(\theta_w)\bigr), \label{eq:tau_box}
\end{align}
where 
$x_I(\theta_w) = \ell_{\text{int}} \sin\theta_w \label{eq:xI}$
denotes the horizontal projection of the torso–exo contact point $I$ with respect to the hip $O$.
The total gravitational torque is
\begin{equation}
\tau_g(\theta_w) \;=\; \tau_w(\theta_w)+\tau_{\text{box}}(\theta_w). \label{eq:tau_g}
\end{equation}
Since the pad is normal to the back, the equilibrium torque generated by the interaction force is 
\begin{equation}
\tau_{\text{int}} \;=\; \ell_{\text{int}}\, F_{\text{int}}. \label{eq:tau_int}
\end{equation}
Balancing torques about $O$ gives
\begin{equation}
\tau_{\text{user}} \;+\; \tau_{\text{int}} \;-\; \tau_g(\theta_w) \;=\; 0, \label{eq:balance}
\end{equation}
with $\tau_{\text{user}}$ the residual muscular torque, required at the equilibrium to compensate for non-ideal balancing $\tau_{\text{int}}$. 
Considering the objective of fully unloading the user (\textit{i.e.} $\tau_{\text{user}}{=}0$), the required interaction torque at the equilibrium results in
\begin{equation}
\tau_{\text{int,eq}}(\theta_w) = \ell_{\text{int}}\,F_{\text{int}}(\theta_w),
\label{eq:assist_torque}
\end{equation}
with $F_{\text{int,eq}}(\theta_w) = \frac{\tau_g(\theta_w)}{\ell_{\text{int}}}$ denoting the corresponding interaction force.
Given that, by design, a residual muscular torque $\tau_{\text{user}}\neq 0$ is always present to preserve user feedback and engagement in the task, to ensure safety by avoiding full overcompensation, and because of inevitable model inaccuracies, the control objective is to provide the user with an assistive torque $\tau_{\text{ass}}$ such that 
\eqref{eq:balance} becomes
\begin{equation}
\tau_{\text{user}} \;+\; \tau_{\text{ass}} \;-\; \tau_g(\theta_w) \;=\; 0,\label{eq:assistance_balance}
\end{equation} such that $\tau_{\text{user}}$ is comfortable and reduced, and $\tau_{\text{ass}}$ is provided timely and consistently with the current task.

\begin{figure*}[t!]
    \centering
    \includegraphics[width=0.99\textwidth]{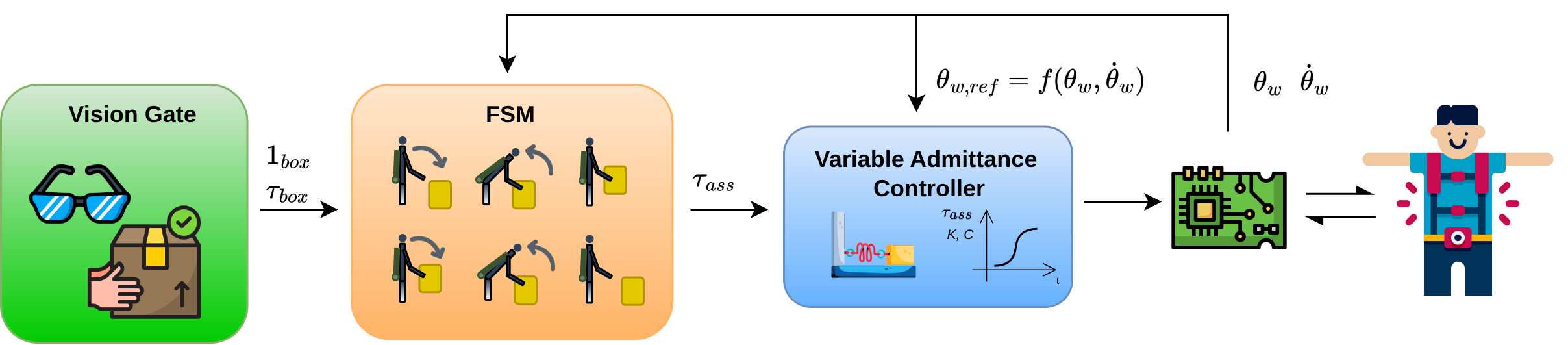}
    \caption{The \dg{\textbf{Vision Gate}} integrates gaze and object detection to identify grasp events. This information is passed to a \dy{\textbf{Finite–State Machine (FSM)}} that governs the assistance logic depending on the user’s posture.
    A \db{\textbf{Variable Admittance Controller}} adapts torque output \(\tau_{ass}\) and impedance parameters (\(K,C\)) according to the current state. An embedded hardware processes the control signal driving the back-support exoskeleton, assisting the user during lifting tasks.}
   \label{fig:ctrl_diagram}
\end{figure*}

\section{Control Methodology}\label{sec:method}
%
The proposed control schema is shown in Figure \ref{fig:ctrl_diagram}.
The perception module (in \dg{\textbf{green}}) monitors the task context (Sec.~\ref{sec:vision}) and triggers state transitions, enabling the exoskeleton to adapt support to the user’s current handling phase.
The finite–state machine (in \dy{\textbf{orange}}) governs when and how the assistance is applied, ensuring smooth scheduling and state–dependent behavior. 
The variable admittance control (in \db{\textbf{blue}}) provides the assistive torque $\tau_{\text{ass}}$ when appropriate, as well as a controlled compliant behavior for the controlled exoskeleton.

\subsection{Assistive Variable Admittance Control}
The lumbar motion is governed by an admittance law defined as:
\begin{equation}
M\,\ddot \theta_w + C\,\dot \theta_w + K\,(\theta_w-\theta_{\text{w,ref}}) 
= \tau_{\text{meas}} + \tau_{\text{ass}}, 
\label{eq:admittance}
\end{equation}
where $\theta_w$ and $\theta_{\text{w,ref}}$ are the measured and reference lumbar angles, respectively; 
$M$, $C$, and $K$ are the virtual inertia, damping, and stiffness parameters; 
$\tau_{\text{meas}}$ denotes the sum of external torques applied about the hip $O$ \footnote{The admittance controller in \eqref{eq:admittance} does not consider the external torque applied by the human only, since the exoskeleton is sensorless. $\tau_{\text{meas}}$ is estimated with motor currents.}; 
and $\tau_{\text{ass}}$ is the assistive torque introduced in Sec.~\ref{sec:exo}.

The reference is incremented by a fixed step $\Delta\theta$ in the direction of motion whenever $|\dot{\theta}w|>\dot{\theta}_{th}$, where $\dot{\theta}_{th}$ denotes a velocity threshold, so that during idle phases $\theta_{\text{w,ref}}\approx\theta_w$, while during motion it is updated. 
This mechanism ensures that the exoskeleton adapts to the user’s voluntary motion while superimposing state–dependent assistance.


The admittance parameters $(C,K)$ are scheduled to provide two distinct interaction modes:
\begin{itemize}
    \item \textit{Soft}: $K{=}0$, $C{=}C_{\text{soft}}$, when the user is idle or moving slowly (backdrivable feel).
    \item \textit{Hard}: $K\!\ge\!K_{\min}$ with $C=2\,\zeta\sqrt{MK}$, when the user initiates faster motion, with constant damping ratio $\zeta$.
\end{itemize}
Mode switches are smoothed by a smoothstep ramp $s(u)=u^2(3-2u)$ of duration $T_K$, and a short hold time $T_{\text{hold}}$ prevents immediate reversion to \textit{Soft}. 
The stiffness update law is
\begin{equation}
K(t) = K_{\text{start}} + \bigl(K_{\text{target}} - K_{\text{start}}\bigr)\,s\!\left(\tfrac{t-t_0}{T_K}\right),
\end{equation}
with $t_0$ the switching instant and $\zeta$ the target damping ratio, and the damping $C$ varies accordingly.

\subsection{Finite–State Assistance Control}
\label{sec:fsm}
For the stooping task, we consider four states:
\[
\mathcal{S}=\Bigl\{
\begin{array}{@{}l@{\quad}l@{}}
0=\texttt{stand\_no\_box}, & 1=\texttt{bend\_to\_pick},\\
2=\texttt{stand\_with\_box}, & 3=\texttt{bend\_to\_place}
\end{array}
\Bigr\}.
\] 
State transitions are defined over $\theta_w$ using two thresholds: 
$\theta_{\text{stand}}$, and $\theta_{\text{bend}}$, corresponding to the upright standing posture, and to the forward bend, respectively. 

The reference assistance is computed as
\begin{equation}
\begin{aligned}
\tau_{\text{ass}}^{\text{ref}}(\theta_w)
&= \gamma \Bigl(\alpha_w\,\tau_w(\theta_w) + \alpha_b\,\tau_{\text{box}}(\theta_w)\Bigr),\\
&\alpha_w,\alpha_b\in\{0,1\},\;\gamma\in(0,1],
\end{aligned}
\label{eq:tau_ass_ref}
\end{equation}
where $\alpha_w$ and $\alpha_b$ select the contributions of the trunk and box torques, respectively, and 
$\gamma$ is a global scaling factor. 
This factor serves two purposes: (i) to avoid replacing the user’s effort entirely, preserving freedom of motion, and (ii) to prevent excessively high torques that could pose risks for both the user and the hardware.
Table~\ref{tab:fsm} summarizes transitions, triggers, and the activation of the weight and box torque contributions. 
In addition to posture thresholds, the table also introduces a perception-derived variable, the vision gate $\mathbf{1}_{\text{box}}$, which can autonomously trigger assistance in place of posture. Its computation and role are detailed in Sec.~\ref{sec:vision}.
 
\begin{table}[t!]
\centering
\caption{FSM transitions and assistance contributions. 
Here $\alpha_w,\alpha_b\in\{0,1\}$ indicate the inclusion of the trunk and box torques, respectively. 
$\alpha_b$ is nonzero only when the vision gate $\mathbf{1}_{\text{box}}$ is active.}
\label{tab:fsm}
\begin{tabular}{p{0.16\columnwidth}p{0.3\columnwidth}p{0.4\columnwidth}}
\hline
\textbf{Transition} & \textbf{Trigger} & \textbf{Contributions} \\
\hline
$0\!\to\!1$ & $\theta_w \ge \theta_{\text{bend}}$ 
& $\alpha_w=0,\;\alpha_b=0$ \\
\hline
$1\!\to\!2$ & $\theta_w \le \theta_{\text{stand}}$ or $\mathbf{1}_{\text{box}} = 1$ 
& $\alpha_w=1,\;\alpha_b=1$ \\
\hline
$2\!\to\!3$ & $\theta_w \ge \theta_{\text{bend}}$ 
& $\alpha_w=0,\;\alpha_b=1$ \\
\hline
$3\!\to\!0$ & $\theta_w \le \theta_{\text{stand}}$ or $\mathbf{1}_{\text{box}} = 0$ 
& $\alpha_w=1,\;\alpha_b=0$ \\
\hline
\end{tabular}
\end{table}

The applied assistance $\tau_{\text{ass}}$ is then obtained by ramping $\tau_{\text{ass}}^{\text{ref}}$ with the previously defined smoothstep profile over a duration $T_{\tau}$:
\begin{equation}
\tau_{\text{ass}}(t)=\tau_{\text{ass}}(t_0)+\bigl(\tau_{\text{ass}}^\text{{ref}}-\tau_{\text{ass}}(t_0)\bigr)\,s\!\left(\tfrac{t-t_0}{T_{\tau}}\right),
\quad t\in[t_0,t_0{+}T_{\tau}].
\end{equation}
This smoothing avoids abrupt discontinuities in the delivered assistance, ensuring a more natural support for the user.

Since posture information alone can be insufficient to determine when assistance is needed, we complement it with a perception layer described next.

\section{Vision-Based Assistance}
\label{sec:vision}

\begin{figure}[b!]
  \centering
  \begin{subfigure}[t]{0.48\columnwidth}
    \includegraphics[width=\linewidth]{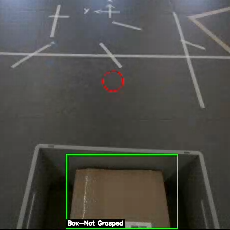}
    \caption{The box is detected in the scene, but the gaze point lies outside all bounding boxes; both $\chi_t^+$ and $\chi_t^-$ are zero.}
  \end{subfigure}\hfill
  \begin{subfigure}[t]{0.48\columnwidth}
    \includegraphics[width=\linewidth]{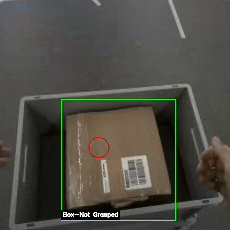}
    \caption{The gaze falls inside a bounding box labeled as \emph{Not Grasped}, setting $\chi_t^- = 1$ while $\chi_t^+=0$.}
  \end{subfigure}

  \vspace{0.6em}

  \begin{subfigure}[t]{0.48\columnwidth}
    \includegraphics[width=\linewidth]{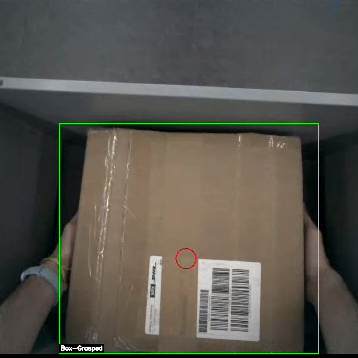}
    \caption{The gaze falls inside a bounding box labeled as \emph{Grasped}, setting $\chi_t^+ = 1$ and switching the vision gate on.}
  \end{subfigure}\hfill
  \begin{subfigure}[t]{0.48\columnwidth}
    \includegraphics[width=\linewidth]{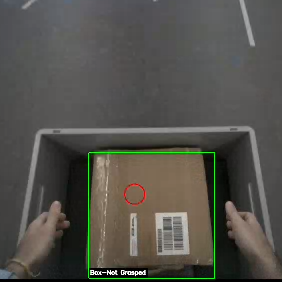}
    \caption{The user realeses the box: The gaze falls inside a bounding box labeled as \emph{Not Grasped}, setting $\chi_t^- = 1$.}
  \end{subfigure}

  \caption{Examples of egocentric frames from the Tobii Pro Glasses 3. Green bounding boxes with labels indicate YOLO detections, while the red circle marks the gaze point. 
  }
  \label{fig:gaze_yolo_examples}
\end{figure}




The perception module combines egocentric video from eye-tracking glasses with gaze coordinates and a fine-tuned YOLO detector \cite{YOLOv9}, enabling real-time discrimination between \textit{grasped} and \textit{not-grasped} boxes. 
This information is converted into the binary gate $\mathbf{1}_{\text{box}}$, which is consumed by the finite--state machine (Sec.~\ref{sec:method}) to compute the additional torque $\tau_{\text{box}}$.


We stream RGB frames from a camera mounted on the glasses together with gaze coordinates $g_x, g_y$, representing the horizontal and vertical fixation point from the binocular eye tracker, as in Figure \ref{fig:gaze_yolo_examples}. 
The gaze is mapped onto the image plane, so that each video frame is paired with the corresponding fixation point.
Object detection is performed by the fine–tuned YOLOv9 described in Sec.~\ref{sec:yolo-finetuning}, 
which outputs bounding boxes $b_i$ with associated class labels 
$c_i \in \{\text{Grasped}, \text{Not Grasped}\}$. 
Given synchronized gaze coordinates $g_x, g_y$, 
we define binary indicators for whether the gaze falls inside at least one bounding box of a given class:
\[
\chi_t^{+} =
\begin{cases}
1 & \text{if } (g_x,g_y)\in b_i \ \text{for some } c_i=\text{Grasped},\\[2pt]
0 & \text{otherwise;}
\end{cases}
\]
\[
\chi_t^{-} =
\begin{cases}
1 & \text{if } (g_x,g_y)\in b_i \ \text{for some } c_i=\text{Not Grasped},\\[2pt]
0 & \text{otherwise.}
\end{cases}
\]
In other words, $\chi_t^+=1$ if the gaze is fixating a box currently labeled as \emph{grasped}, 
while $\chi_t^-=1$ if it falls on a box labeled as \emph{not grasped}. 
These indicators form the basis for the temporal filtering logic that generates 
the final binary gate $\mathbf{1}_{\text{box}}$.

To robustly turn the gate \emph{on}, we require gaze‐on‐Grasped to persist over a time window. 
Let $N_{\text{on}}$ be the time window length and $\rho_{\text{on}}\in(0,1]$ the dwell ratio, \ie the minimum fraction of frames in which gaze must remain on the box.
Defining the sliding average as $r_t^{\text{on}}=\frac{1}{N_{\text{on}}}\sum_{k=0}^{N_{\text{on}}-1}\chi_{t-k}^{+}$, we set $\mathbf{1}_{\text{box}}(t)=1\quad$ if $\quad r_t^{\text{on}}\ge \rho_{\text{on}}$.
Similarly, to turn the gate \emph{off} we monitor gaze‐on‐NotGrasped over a (possibly different) time window $(N_{\text{off}},\rho_{\text{off}})$\footnote{
Using asymmetric windows provides a natural hysteresis that avoids flicker when hands move near the box boundary or when detections are intermittent.}.

The rising edge $\mathbf{1}_{\text{box}}{:}\ 0\!\rightarrow\!1$ authorizes the controller to include the term $\tau_{\text{box}}(\theta_w)$ in the assistance combination (Table~\ref{tab:fsm}), while the falling edge removes it. 
As reported in Table~\ref{tab:fsm}, perception therefore not only enables the $1\!\to\!2$ transition when the gate switches on, but also governs the $3\!\to\!0$ transition when the gate switches off, ensuring timely engagement and disengagement of box-related assistance. 
The temporal filter ensures that these transitions reflect intentional grasp events rather than transient gaze fluctuations.

\section{Experimental Protocol}
\label{sec:experiments}
We conducted a user study to evaluate the proposed vision–gated exoskeleton control during a controlled lifting task. 
The protocol was designed to emulate typical industrial handling scenarios while allowing repeatable measurements across participants. 

\subsection{Task Design}
To test the proposed method, we designed a controlled stooping task for heavy box handling. 
The experiment mimics an industrial scenario in which workers must bend the trunk with limited knee flexion to reach into a container.
A \(4\)\,kg box is placed inside a waist-height bin such that it is not directly reachable with knee flexion alone, requiring participants to bend their torso while keeping their legs nearly extended. 
Each trial consisted of bending to grasp the box, lifting it to a standing posture, and then returning it to the bin and standing free again.

As a vision-enabled gaze-detection device, we used Tobii Pro Glasses~3 in our setup. 
Controller tunings (Sec.~\ref{sec:method}) were fixed across participants: $M{=}0.5$; stiffness scheduled between \textit{Soft} ($K{=}0$) and \textit{Hard} ($K{=}40$) with $T_K{=}0.4$,s; damping from $C{=}2\zeta\sqrt{MK}$ with $\zeta{=}0.9$ ($C_{\text{soft}}{=}5$ when $K{=}0$); assistive torque ramp $T_\tau{=}0.8$,s; lead update step $\Delta\theta_{\text{step}}{=}0.07$,rad, applied only if $|\dot\theta_w|{>}0.2$,rad/s.
The global scaling in~\eqref{eq:tau_ass_ref} was set to $\gamma{=}0.25$ to preserve user effort and comply with hardware torque limits.

\subsection{User-Specific Configuration}
Before the trials, a short calibration was performed to configure the subject-specific thresholds $\theta_{\text{stand}}$ and $ \theta_{\text{bend}}$ used by the finite–state policy (Sec.~\ref{sec:method}). 
Participants wore the exoskeleton without active assistance and executed a few slow bending and standing cycles. 
From encoder data, the following were extracted:
\begin{itemize}
    \item \(\theta_{\text{stand}}\): lumbar angle in upright standing posture,
    \item \(\theta_{\text{bend}}\): lumbar angle at the end of a forward bend.
\end{itemize}
These values were stored and kept fixed throughout the experimental session.

The subjective parameters $m_w$ and $l_w$ are adjusted according to each user's parameters, following biomechanical models presented in  \cite{de1996adjustments, plagenhoef1983anatomical}.
In particular, we compute the weight of the upper part of the body $\approx55\%$ of the total weight, and the upper body center of mass ($W$ in Figure \ref{fig:forze-coppie}) coincident with the sternum.
Finally, according to each user's height, mechanical adjustments of the exoskeleton are allowed, changing $l_{int}$.

\subsection{Experimental Conditions}
After calibration, participants performed the stooping task under three conditions:
\begin{enumerate}
    \item \textbf{EC1 – No exoskeleton}: baseline trials without wearing the device.
    \item \textbf{EC2 – Exoskeleton without vision}: assistance provided by the finite–state controller based only on posture thresholds.
    \item \textbf{EC3 – Exoskeleton with vision}: vision–gated assistance integrating gaze–based grasp awareness.
\end{enumerate}
In EC2 and EC3, participants were not informed about vision activation to avoid bias.

\subsection{Participants and Trial Structure}
A total of 15 healthy volunteers participated in the study
(mean age $28.1 \pm 4.3$ years, mean height $178.7 \pm 6.4$\,cm, mean weight $74.0 \pm 7.8$\,kg). 
Each subject performed five consecutive stooping task repetitions per condition, yielding \(15\times 5\times 3=225\) trials overall. Short rest pauses were allowed between conditions to mitigate fatigue. 

At the end of the experimental session, participants completed a 13–item questionnaire adapted from~\cite{questionnaire}, rated on a 4–point Likert scale (1 = strongly disagree, 4 = strongly agree).
Items probed physical demand, fluency of the human–robot team, perceived contribution of the robot, trust, comfort, and mutual understanding, as detailed in Table~\ref{tab:questionnaire}. 
The first item (“The lifting task was not physically demanding”) was also asked in the baseline condition without the exoskeleton.

\begin{table}[t!]
\centering
\caption{Post–trial questionnaire (4–point Likert scale). The first item was asked in all conditions, while items Q2–Q13 were asked only when the exoskeleton was worn.}
\label{tab:questionnaire}
\begin{tabular}{p{0.05\columnwidth}p{0.85\columnwidth}}
\hline
 & \textbf{Question} \\
\hline
Q1 & The lifting task was not physically demanding. \\
Q2 & The human–robot team worked fluently together. \\
Q3 & The robot contributed to the fluency of the interaction. \\
Q4 & I didn’t have to carry the weight to improve the human–robot team. \\
Q5 & The robot contributed equally to the team performance. \\
Q6 & The robot was an important team member on the team. \\
Q7 & I trusted the robot to do the right thing at the right time. \\
Q8 & The robot was intelligent. \\
Q9 & The robot was trustworthy. \\
Q10 & The robot was committed to the task. \\
Q11 & I feel comfortable with the robot. \\
Q12 & The robot and I understand each other. \\
Q13 & The robot perceives accurately what my goals are. \\
\hline
\end{tabular}
\end{table}

\section{Results}
\label{sec:results}

\subsection{Perception Performances}
\label{sec:yolo-finetuning}
We fine-tune YOLOv9 to discriminate between two visual states of the target object:
\emph{Box-Grasped} and \emph{Box-Not Grasped}.
While the grasp state alone indicates whether the object is currently held, combining it with posture information allows the system to infer the current phase of the pick-and-place task (\textit{e.g.}, bending to pick vs.\ standing with the box).
This visual cue acts as the binary gate $\mathbf{1}_{\text{box}}$ in the finite–state controller of Figure~\ref{fig:ctrl_diagram}, which regulates whether the box–related torque contribution $\tau_{\text{box}}$ should be included in the delivered assistance.

The dataset contains \textit{1651} images with \textit{1982} annotated boxes spanning two labels: \textit{box-grasped} (\(622\)) and \textit{box-not grasped} (\(1{,}360\)); seven box types are represented. 

 
Since gaze tracking was calibrated individually for each participant, gaze remained consistently aligned with the manipulated box during the task. 
Thus, the detection accuracy of YOLO can be directly interpreted as the performance of the full vision-gated trigger. 
Table~\ref{table:visionmetric} summarizes the resulting precision, recall, and F1, showing high accuracy on both classes.

\begin{table}[t!]
    \centering
    \caption{Evaluation metrics of the vision–based grasp detection module.}
    \begin{tabular}{| c | c | c | c |}
    \hline
    \textbf{Label} & \textbf{Precision} & \textbf{Recall} & \textbf{F1-Score} \\
    \hline
     \textbf{Box-Grasped} & 96.12\% & 93.71\% & 94.90\% \\
    \hline
     \textbf{Box-Not Grasped} & 90.54\% & 90.54\% & 90.54\% \\
    \hline
     \textbf{All (macro avg)} & 92.84\% & 91.86\% & 92.35\% \\
    \hline
    \end{tabular}
    \label{table:visionmetric}
\end{table}

\begin{figure}[b!]
    \centering
    \includegraphics[width=0.99\linewidth]{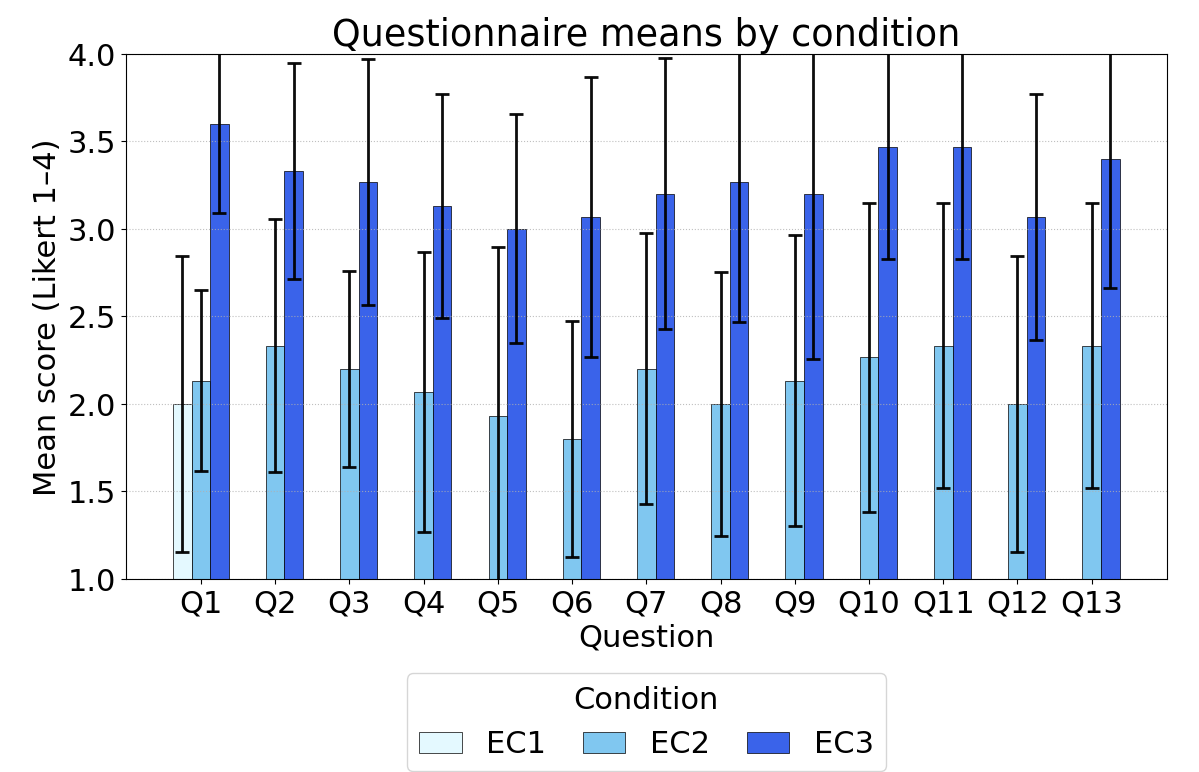}
    \caption{Mean questionnaire scores under the three experimental conditions. Error bars show standard deviations.}
    \label{fig:means}
\end{figure}

\subsection{Questionnaire}

\begin{table}[t!]
\centering
\caption{Descriptive statistics (mean $\pm$ SD, min--max) per question and condition.}
\label{tab:desc}
\begin{tabular}{lccc}
\hline
 & EC1 & EC2 & EC3 \\
\hline
Q1  & 2.00$\pm$0.85 [1--4] & 2.13$\pm$0.52 [1--3] & 3.60$\pm$0.51 [3--4] \\
Q2  & --                  & 2.33$\pm$0.72 [1--4] & 3.33$\pm$0.62 [2--4] \\
Q3  & --                  & 2.20$\pm$0.56 [1--3] & 3.27$\pm$0.70 [2--4] \\
Q4  & --                  & 2.07$\pm$0.80 [1--4] & 3.13$\pm$0.64 [2--4] \\
Q5  & --                  & 1.93$\pm$0.96 [1--4] & 3.00$\pm$0.65 [1--4] \\
Q6  & --                  & 1.80$\pm$0.68 [1--3] & 3.07$\pm$0.80 [2--4] \\
Q7  & --                  & 2.20$\pm$0.77 [1--4] & 3.20$\pm$0.77 [2--4] \\
Q8  & --                  & 2.00$\pm$0.76 [1--3] & 3.27$\pm$0.80 [2--4] \\
Q9  & --                  & 2.13$\pm$0.83 [1--3] & 3.20$\pm$0.94 [1--4] \\
Q10 & --                  & 2.27$\pm$0.88 [1--4] & 3.47$\pm$0.64 [2--4] \\
Q11 & --                  & 2.33$\pm$0.82 [1--4] & 3.47$\pm$0.64 [2--4] \\
Q12 & --                  & 2.00$\pm$0.85 [1--3] & 3.07$\pm$0.70 [2--4] \\
Q13 & --                  & 2.33$\pm$0.82 [1--4] & 3.40$\pm$0.74 [2--4] \\
\hline
\end{tabular}
\end{table}

To determine whether differences among conditions are statistically meaningful, we selected non–parametric tests appropriate for ordinal Likert-scale data with repeated measurements per subject. Specifically, for Q1 (the only question asked in all three conditions), we used the Friedman test (omnibus test for three related samples), with Kendall’s \(W\) as effect size. For comparisons between two conditions (EC2 vs EC3) on Q2–Q13, we used paired Wilcoxon signed–rank tests, with Holm correction for multiple comparisons, and reported rank-biserial correlations as effect sizes.

\subsubsection{Descriptive Statistics}

Figure~\ref{fig:means} shows mean questionnaire scores per item across the three experimental conditions, with standard deviations as error bars. Table~\ref{tab:desc} provides means, standard deviations, and min-/max values per condition and question.

Key descriptive results include:
\begin{itemize}
    \item In the baseline (EC1) condition (Q1 only), participants rated the task as physically demanding, mean = 2.00 \(\pm\) 0.85 (range 1.0–4.0).
    \item With exoskeleton but no vision gating (EC2), Q1 increased slightly to 2.13 \(\pm\) 0.52.
    \item The largest improvement was with vision gating (EC3): Q1 = 3.60 \(\pm\) 0.51 (range 3.0–4.0).
    \item For Q2–Q13 (only under EC2 and EC3), mean scores under EC2 ranged approximately between ~1.80 and ~2.33 across questions; under EC3 they ranged between ~3.00 and ~3.60 (see Table~\ref{tab:desc}).
\end{itemize}

\subsubsection{Inferential Statistics}
Friedman test, applied to Q1, revealed a significant effect of condition on perceived physical demand, \(\chi^2(2)=22.571\), \(p < 0.001\), Kendall’s \(W=0.752\) (with \(N=15\) subjects).  
Post-hoc Wilcoxon signed-rank tests with Holm correction showed that EC3 is significantly better than EC1 (\(p_{\text{holm}}=0.0025\)) and also better than EC2 (\(p_{\text{holm}}=0.0020\)), while the difference between EC1 and EC2 did not reach significance (\(p=0.527\)).

For Q2–Q13, Wilcoxon signed--rank tests with Holm correction indicated that the vision--gated condition significantly outperformed the posture--only condition across all items. 
All questions remained significant after correction ($p_{\text{holm}}{<}0.05$), with the smallest $p_{\text{holm}} \approx 0.032$ (Q8) and the largest $p_{\text{holm}} \approx 0.045$. 
Rank--biserial effect sizes were large across questions (typical range $r\in[0.67,1.00]$).

\subsubsection{Summary}

\begin{itemize}
  \item Vision-gated exoskeleton assistance significantly reduces perceived physical demand compared to both no device and posture-based assistance (large effect on Q1).
  \item Across questions addressing fluency, trust, comfort, and mutual understanding, participants consistently rated the vision condition higher; all items remained significant after Holm correction.
  \item Effect sizes were large overall; for Q1 Kendall’s \(W=0.752\) indicates a substantial effect. For Q2–Q13, rank-biserial correlations were consistently high (often $r\geq0.80$), with unanimous improvements on Q4.
\end{itemize}

These results strongly support that integrating vision for grasp detection improves subjective experience across multiple dimensions in physically demanding tasks.

\begin{figure*}[t!]
    \centering
    \includegraphics[width=0.95\textwidth]{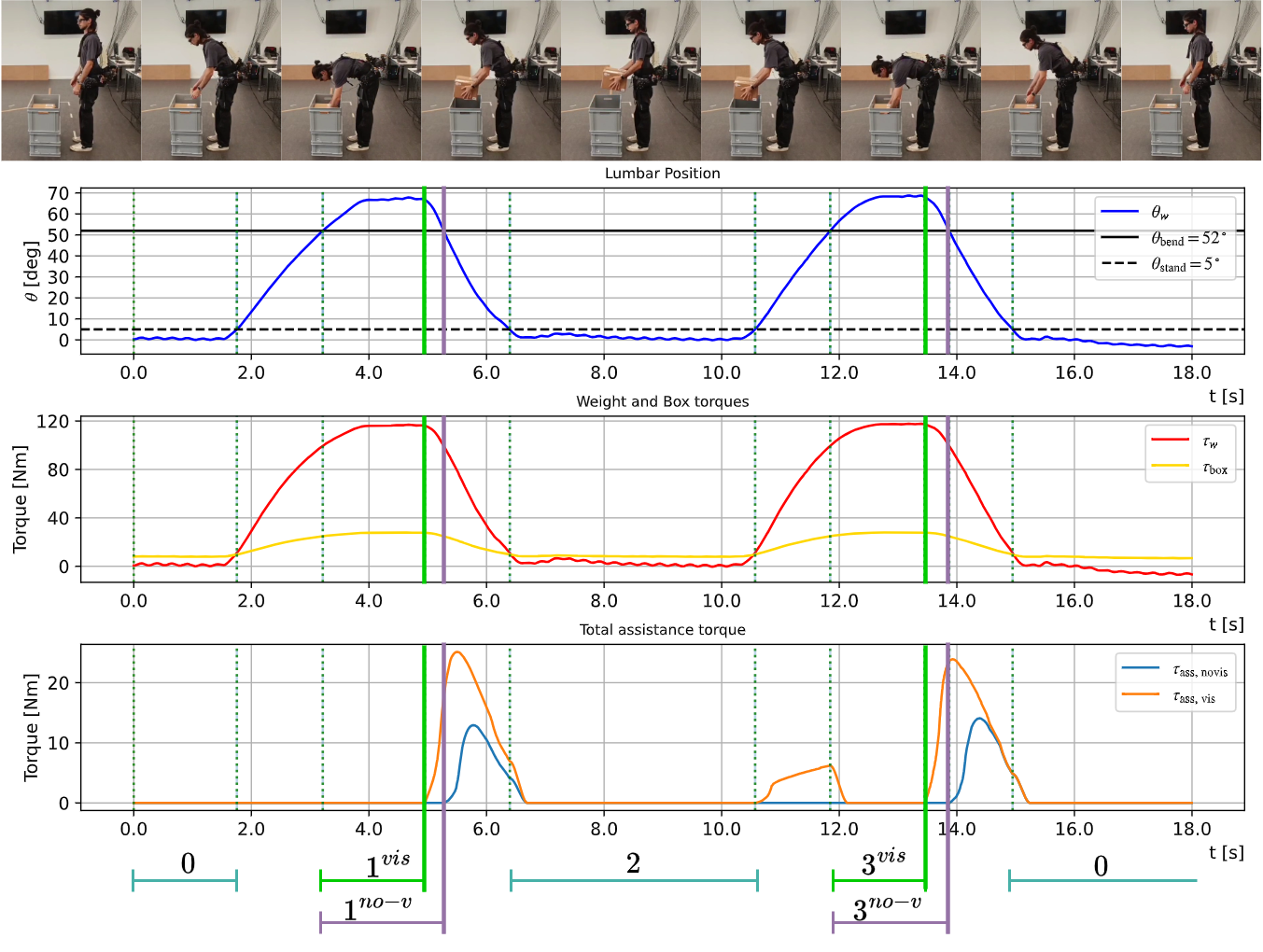}
    \caption{Example stooping task performed with vision--gated assistance. First Row: sequence of frames of the user during the execution of the task. Second Row: lumbar angle trajectory with reference thresholds (black continuous and dashed lines). Third Row: estimated torques $\tau_w$ and $\tau_{box}$. Fourth Row: total assistance torque with vision and without vision. For comparison, the posture--based assistance ($\tau_{\text{ass,novis}}$) is shown aligned to the same motion.}
    \label{fig:stooping_example}
\end{figure*}

\subsection{Discussion}
Participants perceived a clear difference between the two assistance modes, despite being unaware of whether vision was enabled. 
Without vision, some subjects noted a lag in assistance onset, especially at the critical transition from bending to lifting, where the controller had to infer the user’s intent from posture change alone. 
As illustrated in Figure~\ref{fig:stooping_example}, vision--based gating led to shorter state~1 phases compared to posture--only control, reducing the lag between the actual grasp and the onset of assistance, and resulting in greater perceived responsiveness. 
Consequently, participants did not need to exert additional effort to reach the $\theta_{\text{bend}}$ threshold before receiving support, as the controller started assisting already at the moment of grasp. 
This anticipatory action is visible in the total assistance torque, where $\tau_{\text{ass,vis}}$ rises earlier and provides greater support during the transition from bending to lifting. 
Finally, with vision gating, the assistance included the additional $\tau_{\text{box}}$ term, leading to higher overall torque and more effective compensation during both the 1$\rightarrow$2 and 2$\rightarrow$3 transitions.
These subjective impressions align with the questionnaire results, which showed significantly higher ratings of fluency, trust, and comfort with vision-based gating. 
Vision allowed the controller to anticipate task context, increasing the perceived responsiveness and intelligence of the system.

A relevant limitation of the proposed study is that it considers a single box type with known weight ($4$\,kg) under controlled laboratory conditions. 
This allows modeling $\tau_{\text{box}}$ directly, but may not hold in realistic scenarios where payloads vary. 
This limitation can be overcome by incorporating online weight estimation techniques, either through perception (\textit{e.g.}, label or barcode reading, vision-based size inference) or through dynamic adaptation using torque feedback. 

\section{Conclusions}
\label{sec:conc}
We presented a vision–gated control strategy for a lumbar exoskeleton, combining a finite–state task policy with variable admittance control and perception–based assistance. 
A YOLOv9 detector, fine–tuned to discriminate grasped and not–grasped boxes, was integrated with binocular gaze tracking from Tobii Pro Glasses 3 to provide a binary gate for activating load–specific support. 

Experiments with 15 participants performing repeated stooping tasks demonstrated that vision–based gating significantly improved subjective ratings of physical demand, fluency, trust, and comfort compared to both unassisted lifting and exoskeleton use without vision. 
Participants reported that assistance was more timely and smoother, particularly during the critical transitions of lifting and placing a load, highlighting the potential of combining lightweight vision and gaze sensing with model–based control to achieve more intuitive and responsive human–robot collaboration. 

Future work will extend the evaluation to more varied loads and tasks, and explore physiological metrics alongside subjective measures to better quantify the benefits of perception–gated assistance. 
In addition, we aim to investigate alternative sensing modalities such as wearable body cameras or microphones, and to enhance the vision pipeline to recognize different object types and weights, enabling adaptive assistance strategies tailored to the specific item being handled. 
Finally, in the presented study, the exoskeleton leg motors are not controlled. 
Future works will address the integrated and coordinated control of the lumbar and hip joints, allowing assistance for different handling tasks (\textit{e.g.}, squatting and walking).


\bibliographystyle{IEEEtran}
\bibliography{bibliography.bib}

\end{document}